\documentclass[letterpaper, 10 pt, conference]{ieeeconf}  
\IEEEoverridecommandlockouts     
\usepackage{graphicx}      

\newtheorem{remark}{Remark}

\usepackage{hyperref}
\usepackage{amsfonts}
\DeclareMathSymbol{\shortminus}{\mathbin}{AMSa}{"39}
\let\labelindent\relax
\usepackage{enumitem}
\usepackage{multirow}
\usepackage{flushend}
\usepackage{tikz}
\usepackage{booktabs}
\usepackage{array}
\usepackage{stfloats}
\usepackage{float}
\usepackage{amsmath}
\usepackage{epsfig}
\usepackage{arydshln}
\usepackage{algpseudocode}
\usepackage{verbatim}
\usepackage{enumerate}
\usepackage{rotating}
\usepackage{color}
\usepackage{flushend}
\usepackage{mathrsfs}
\usepackage{amssymb}
\usepackage{placeins}
\usepackage[ruled,vlined,linesnumbered]{algorithm2e}
\usepackage{cite}
\usepackage{float}
\usepackage{soul}
\usepackage{graphicx}
\usepackage[caption=false,font=footnotesize]{subfig} 
\usepackage{fancyhdr}
\usetikzlibrary{arrows.meta}
\usepackage{url}
\usepackage{makecell} 
\usepackage{mathtools}
\usepackage[font=small,labelsep=space,labelsep=colon]{caption}
\usepackage{amsmath} 
\usepackage[para,online,flushleft]{threeparttable}
\definecolor{myyellow}{RGB}{205,192,176} 
\definecolor{mygray}{RGB}{192,192,192}
\definecolor{mycyran}{RGB}{75,211,239}

\def \x{{\mathbf{x}}}

\def \until{{\mathbf{U}}}
\def \eventually{{\mathbf{F}}}
\def \always{{\mathbf{G}}}

\def \RR{\mathbb{R}}
\def \path{\textsf{Path}}

\def \RR{\mathbb{R}}

\title{\bf Bridging Perception and Planning: Towards End-to-End Planning\\ for Signal Temporal Logic Tasks} 

\author{Bowen Ye, Junyue Huang, Yang Liu, Xiaozhen Qiao and Xiang Yin
\thanks{This work was supported by the National Science and Technology Major Project (2025ZD1600701-5) and the National Natural Science Foundation of China (62573291,62533017,62173226).}
 \thanks{B. Ye, J. Huang and X. Yin are with the School of Automation \& Intelligent Sensing, Shanghai Jiao Tong University, Shanghai 200240, China.
	{\tt\small  E-mail: \{yebowen1025, hjy-564904993, yinxiang\}@sjtu.edu.cn}. Y. Liu is with the University of Minnesota, Twin Cities.
{\tt\small  E-mail: \{liu03222\}@umn.edu}. X. Qiao is with the School of Information Science and Technology, University of Science and Technology of China, Hefei, 230026, China.
\tt\small{E-mail:\{xiaozhennnqiao\}@mail.ustc.edu.cn}
}
}

\begin{document}
\maketitle 
\begin{abstract}
We investigate the task and motion planning problem for Signal Temporal Logic (STL) specifications in robotics. Existing STL methods rely on pre-defined maps or mobility representations, which are ineffective in unstructured real-world environments.
We propose the \emph{Structured-MoE STL Planner} (\textbf{S-MSP}), a differentiable framework that maps synchronized multi-view camera observations and an STL specification directly to a feasible trajectory. S-MSP integrates STL constraints within a unified pipeline, trained with a composite loss that combines trajectory reconstruction and STL robustness. A \emph{structure-aware} Mixture-of-Experts (MoE) model enables horizon-aware specialization by projecting sub-tasks into temporally anchored embeddings.
We evaluate S-MSP using a high-fidelity simulation of factory-logistics scenarios with temporally constrained tasks. Experiments show that S-MSP outperforms single-expert baselines in STL satisfaction and trajectory feasibility. A rule-based \emph{safety filter} at inference improves physical executability without compromising logical correctness, showcasing the practicality of the approach.
\end{abstract}

\section{Introduction} 
Signal Temporal Logic (STL) has emerged as a widely-used specification language in the design of autonomous systems, primarily due to its quantitative robustness semantics and the wealth of temporal operators it offers. It has been successfully applied across a broad range of engineering applications \cite{belta2019formal,yin2024formal,zhao2025no,yu2026signal}. Particularly, in the field of robotics motion and task planning, STL has garnered significant attention  \cite{fossdal2024past,yu2024model,wang2024sleep,dietrich2025symbolic}.

Despite its success in robotic task and motion planning, current STL methods remain largely restricted to scenarios where structured system information is available \emph{a priori}. This structured information typically includes:
(i) environmental semantics (e.g., atomic predicates),
(ii) the abstracted workspace for discrete-level planning (commonly represented as  grids or polytopic regions), and
(iii) system dynamics when planning over both tasks and motions.
Most existing STL planning approaches, including encoding-based \cite{raman2014model,takano2021continuous}, gradient-based~\cite{gilpin2020smooth,leung2023backpropagation,meng2023signal} and learning-based optimization methods~\cite{cho2018learning,wang2024synthesis,wang2024tractable,guo2024temporal,liu2025zero}, fall into this category, as they all fundamentally rely on predefined structural information at some level.

Current methods typically separate environmental perception, structured information modeling, and STL task planning into distinct stages. Consequently, most efforts emphasize planning over pre-structured abstractions, while the influence of perceptual uncertainty and abstraction quality from unstructured observations is comparatively underexplored. In practice, structured information must be inferred from nonlinear system dynamics and high-dimensional sensory inputs (e.g., images), yet conventional pipelines do not support direct reasoning over such raw data, leading to brittle performance under distribution shift or missing map annotations. An end-to-end framework that learns to satisfy STL specifications directly from perceptual inputs can reduce manual discretization and better couple perception with decision-making, consistent with recent successes of end-to-end learning in robotics.

Building on this perspective, recent studies in autonomous driving~\cite{hu2023planning, jia2025drivetransformer} and robotic manipulation empowered by vision–language–action (VLA) models~\cite{black2024pi0, zitkovich2023rt} show that end-to-end learning paradigms are advantageous in unifying representation and decision-making. Inspired by these findings, we believe that such unification could significantly enhance STL-constrained planning. However, solving STL tasks \emph{directly} from raw visual observations remains an unexplored challenge. Existing methods still rely on explicit map abstractions or low-dimensional state inputs. This gap motivates our pursuit of a truly end-to-end framework that integrates perception, temporal logic reasoning, and trajectory generation into a single differentiable architecture.

In this work, we propose an end-to-end autoregressive transformer framework that ingests synchronized multi-view camera streams along with an STL specification, and outputs a physically feasible trajectory guaranteed to satisfy the prescribed temporal-logic constraints. However, two key challenges impede the realization of such an end-to-end architecture. First is the issue of dataset unavailability. To our knowledge, no public corpus offers synchronized multi-view imagery, formal STL task specifications, and expert trajectories simultaneously, complicating supervised training and objective benchmarking. Second is the complexity of spatio-temporal learning. Specifically, the rich temporal operators and spatial predicates of STL create a highly non-convex, long-horizon optimization landscape, making gradient-based learning more demanding than in conventional trajectory-prediction settings.
To address these challenges, our work makes the following contributions:

\begin{itemize}
\item \textbf{End-to-End STL Planner.}
We present the first baseline that performs STL-constrained trajectory synthesis \emph{directly} from synchronized multi-view camera observations. A single autoregressive Transformer generates a physically feasible trajectory satisfying any given STL specification, without requiring intermediate map abstractions. 
\item \textbf{Structure-Aware Mixture of Experts.}  
We introduce a structure-aware MoE model that decomposes each STL formula into temporally anchored sub-task tokens and routes them to operator-specific experts. This facilitates efficient end-to-end learning of STL-compliant trajectories and enhances logical reasoning by exploiting STL compositionality.

\item \textbf{Benchmark Dataset and Empirical Validation.}  
We construct a high-fidelity, Gazebo-based benchmark that combines temporally synchronized multi-view camera streams with formally annotated STL specifications and expert trajectories across a range of factory-logistics scenarios. This is the first publicly available corpus designed for supervised, end-to-end STL learning. We then demonstrate that the proposed model achieves state-of-the-art STL satisfaction and trajectory feasibility, with improved performance and without incurring additional planning latency compared to a single-expert transformer baseline. 
\end{itemize} 

\section{Related Works}

\textbf{End-to-End Model Paradigm.} \textsc{UniAD}~\cite{hu2023planning} pioneered the end-to-end framework in autonomous driving, unifying multiple perception tasks into a single architecture to directly support downstream planning, enhancing planning quality and efficiency. End-to-end designs have since been explored in autonomous driving and robotic manipulation. In autonomous driving, multi-task BEV/Vector transformers and closed-loop differentiable planners unify perception, prediction, and planning~\cite{jiang2023vad, pan2024vlp, ye2025}, while diffusion-based trajectory generators produce multimodal futures and safety-guided sampling~\cite{jiang2025transdiffuser, liao2025diffusiondrive}. In manipulation, diffusion policies in joint or SE(3) spaces demonstrate strong long-horizon performance~\cite{chi2023diffusion}, and VLA models align semantics with control via instruction grounding and policy distillation~\cite{brohan2022rt, leal2024sara}. These advancements highlight the advantages of end-to-end approaches, such as better coupling of representation and decision-making, uncertainty-aware generative modeling, and language-conditioned grounding. Building on this, we introduce the first end-to-end solution for solving STL-specified tasks using raw sensory inputs and an STL specification directly.

\textbf{Mixture-of-Experts Models.} Sparse MoE improves model capacity by replacing dense layers with a bank of experts, activated conditionally via a learned router, enabling task specialization and computational efficiency. In large language models, MoE reduces perplexity and improves multitask accuracy without increasing inference costs~\cite{dai2024deepseekmoe}. In autonomous driving, expert routing disentangles maneuver styles and scenario primitives~\cite{yang2025drivemoe}, while in robotics, MoE handles task heterogeneity and long-tail data~\cite{huang2024mentor}. Conditional routing directs rare or context-specific tasks to specialized experts, enhancing performance, data efficiency, and generalization. Motivated by these findings, we integrate a structure-aware MoE into our model, adapting routing and expert designs to align with the compositional and temporal nature of STL tasks.

\section{Signal Temporal Logic Specifications}\label{sec:pro} 
Let $X$ denote the state space and let $\mathbf{x}_{0:T}=(x_0,\ldots,x_T)\in  X^T$ be a finite trajectory. 
The syntax of Signal Temporal Logic (STL) is given by
\begin{equation}
    \phi ::= \top\mid\pi^{\mu}\mid\neg\phi\mid\phi_{1}\wedge\phi_{2}\mid\phi_{1}\until_{[a,b]}\phi_{2}
\end{equation}
where $\top$ is the true predict, $\pi^{\mu}$ is a predicate whose truth value is determined by the sign of its underlying predicate
function $\mu : \RR^{n}\to\RR$ and is true if $\mu(x_k)\geq 0$; otherwise it is false. Notation $\neg$ and $\wedge$ are the standard Boolean operators ``negation'' and ``conjunction'', respectively, which
can further induce ``disjunction'' $\vee$ and ``implication'' $\rightarrow$. Notation $\until_{[a,b]}$ is the temporal operator ``until", where $a, b \in\RR_{\geq0}$ are the time instants. 
STL formulae are evaluated on state sequences. 
We use $(\x,t)\models\phi$ to denote that the sequence $\x$ satisfies STL formula $\phi$ at instant $t$. The semantics of STL are given by: 
$(\x,t)\models \phi_1\until\phi_2$    iff   $\exists t'\in[t+a,t+b]:(\x,t')\models\phi_2,\forall t''\in[t+a,t']:(\x,t'')\models\phi_1$. The reader is referred to \cite{maler2004monitoring} for more details on the semantics of STL. 
Furthermore, we also induce two important temporal operators ``eventually" and ``always" by $\eventually_{[a,b]}\phi:=\top\until_{[a,b]}\phi$ and $\always_{[a,b]}:=\neg\eventually_{[a,b]}\neg\phi$, respectively. For a given sequence $\x$, we write $\x\models\phi$ whenever $(\x,0)\models\phi$.

Apart from the syntax and semantics of STL, we also briefly review the quantitative semantics of STL as follows. Given a signal $\x$ and time instant $t$, the robustness semantics of $\x$ at $t$ is recursively defined as follows:
$\rho^\mu(\x,t)\!=\!\mu(x_t)$; $\rho^{\neg\mu}(\x,t)\!=\!-\mu(x_t)$; $\rho^{\phi_1\wedge\phi_2}\!=\!\min(\rho^{\phi_1}(\x,t),\rho^{\phi_2}(\x,t))$;  $\rho^{\always_{[a,b]}\phi }(\x,t)\!=\!\min_{t'\in[t+a,t+b]}\left(\rho^\phi(\x,t')\right)$; $\rho^{\phi_1\until\phi_2}(\x,t)\!=\!\max_{t'\in[t+a,t+b]}\left(\min(\rho^{\phi_2}(\x,t')),\min_{t''\in[t,t']}\rho^{\phi_1}(\x,t'')\right)$. Naturally, for a signal $\x$, we have $\x\models\phi\Leftrightarrow\rho^\phi(\x,0)>0$.

Following~\cite{8062628}, we adopt a smooth surrogate of the robustness to enable end-to-end training. 
Because $\min/\max$ and $\inf/\sup$ introduce non-differentiable truncations and flat regions, direct backpropagation through $\rho^\phi$ is unstable. 
We therefore replace $\max$ $(\sup)$ and $\min$ $(\inf)$ operators with temperature-controlled smooth counterparts $\max/\min$ :
\begin{subequations}
    \begin{align}
        & \widetilde{\max}_{k}(x_1, x_2,\dots):=\frac{1}{k}\log(e^{kx_1} + e^{kx_2} + \dots) \\
        & \widetilde{\min}_{k}(x_1, x_2,\dots):= -\widetilde{\max}_{k}(-x_1, -x_2,\dots)
    \end{align}
\end{subequations}
where k is a scaling factor for this approximation. If $k\to\infty$, the operator $\widetilde{\max} = \max$ and similarly $\widetilde{\min} = \min$. We use $k = 300$ in our training, the ablation study on $k$ is in Table~\ref{tab:scaling-law}.

As the general STL is highly expressive, its hypothesis class can be overly rich relative to the available data, which makes it difficult for a learning-based end-to-end model to recover precise semantics and yields unstable gradients during training. To ensure learnability and tractable monitoring, we restrict attention to a widely used finite-horizon fragment that still captures the majority of robotic tasks.
Specifically,  we adopt a discrete-time semantics with horizon \(T\in\mathbb{N}\) and timestamps \(t\in\{0,1,\dots,T\}\).
All temporal operators are \emph{bounded} in time, with integer bounds \(0\le a\le b\le T\).
We further limit temporal nesting depth to \(1\): each temporal operator applies directly to a temporal-free Boolean formula.
Formally, let atomic predicates be \(\mu(x_t)\) with robustness \(\rho_\mu[t]\in\mathbb{R}\), and let \(\chi\) denote any Boolean combination of atoms using \(\wedge,\vee\) (no temporal operators inside).
Our fragment \(\mathsf{STL}_{\mathrm{B,depth}\le1}\) is generated by
\[
\varphi \;::=\; \chi \;\mid\; \eventually_{[a,b]}\,\chi \;\mid\; \always_{[a,b]}\,\chi \;\mid\; \varphi_1\wedge\varphi_2 \;\mid\; \varphi_1\vee\varphi_2
\]
This fragment covers canonical reach-avoid, liveness (\(\eventually\)), and safety (\(\always\)) specifications frequently used in robotics, while keeping monitoring and gradient-based training tractable.

\begin{remark}
    We temporarily exclude ``implication" (\(\rightarrow\)), ``negation" (\(\lnot\)), and the ``until" operator ($\until$).
Negation satisfies \(\rho_{\lnot\varphi}[t] = -\,\rho_\varphi[t]\), which flips the objective’s sign and amplifies gradient-direction uncertainty under compositional nesting; implication reduces to \(\lnot\varphi \vee \psi\) and inherits the same pathology.
For Until, the discrete-time robustness takes a nested min–max form:
\[
\rho_{\varphi\,\until_{[a,b]}\,\psi}[t]
=\!\!\! \max_{\tau\in\{a,\dots,b\}}
\!\!\Big(\!\min\!\Big(\rho_\psi[t+\tau],\!\! \min_{s\in\{0,\dots,\tau-1\}}\rho_\varphi[t+s]\Big)\!\Big)
\]
which introduces deep non-smooth min–max structure and strong nonlocal temporal coupling, yielding ill-conditioned objectives and highly irregular (Clarke) subgradients that impede stable learning.
\end{remark}

\section{Our Scenario and Problem Formulation}
Our overall objective is to design an end-to-end STL planner whose inputs are:
\begin{itemize}
\item raw multi-view camera data of the workspace, without any preprocessing such as semantic recognition or region discretization; and
\item a structural STL formula that the agent needs to fulfill.
\end{itemize}
The output is a feasible trajectory for the ego-agent, ensuring that it both satisfies the STL task and is feasible for the agent's dynamics. The trajectory is represented as a sequence of waypoints $\tau = {(x_t, y_t, \psi_t)}_{t=1}^{T_f}$, where $T_f$ is the planning horizon, and $(x_t, y_t, \psi_t)$ denotes the position and heading of the agent at time $t$, expressed in a fixed world (absolute) reference frame.

During the training phase, we assume the availability of a dataset containing task-related trajectories with raw multi-view environment RGB data. Specifically, each data entry includes synchronized multi-view RGB streams with calibrated intrinsics and extrinsics, an STL formula, and an expert reference trajectory. Particularly, we make the following  assumptions for training data and execution environment
\begin{itemize}
    \item 
    \textbf{Consistent environmental semantics:} 
    The semantic information of the environment remains unchanged, and the training data includes all possible image semantics that the agent may encounter during execution.
    \item 
    \textbf{Labeled target areas:}
    To enable the model to learn the correspondence between target regions and STL task specifications from data, we mark different regions with distinct colors. Note that the model input consists only of multi-view concentrated image observations, without any explicit geometric annotations (e.g., region coordinates or masks).
    \item 
    \textbf{Consistent system dynamics:} 
    The dynamics of the system in both the training and execution environments are assumed to be the same.
    \item 
    \textbf{Complete multi-view coverage:}  The multi-view camera setup is assumed to fully cover the operating scene, with no blind spots. For example, in autonomous driving, this could be achieved using cameras mounted at various points around the vehicle. In our factory scenario example, multiple cameras suspended from different positions on the ceiling may achieve full coverage.
\end{itemize}
Furthermore, without loss of generality, we assume that the training and execution environments share the same physical dimensions. However, the positions of task zones and obstacle regions are allowed to vary, thereby generating diverse scenarios.

\begin{figure*}[!t]
  \centering
  \includegraphics[width=0.90\textwidth,trim={0 0mm 0 0mm},clip]{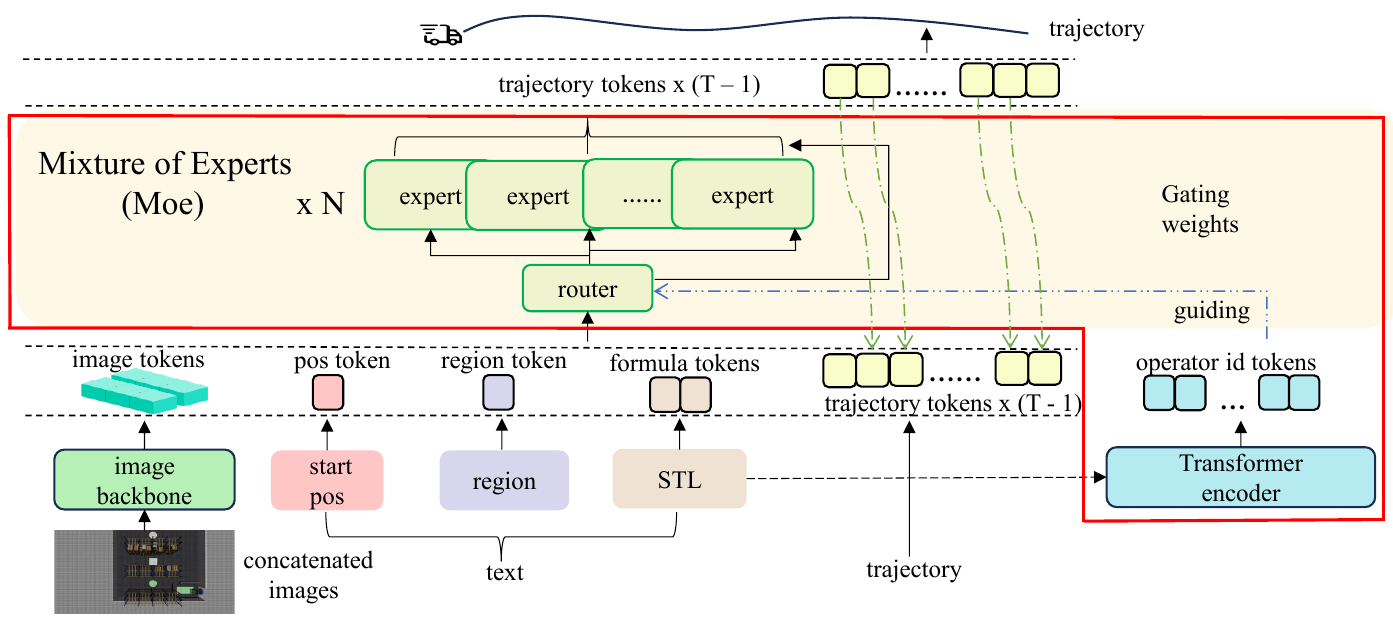}
  \caption{The Overall Architecture of Model.}
  \label{fig:model}
\end{figure*}

\section{E2E Structural Moe Model}\label{sec:model}
The overall pipeline of \textbf{S-MSP} is shown in Fig~\ref{fig:model}, which is a Transformer-based, end-to-end planner that maps multi bird-eye-view images and a Signal Temporal Logic (STL) task to a future waypoint sequence. The \textbf{S-MSP} model fuses four conditioning pathways: vision, STL text, pose/region context, and a \emph{structural} encoding of the STL formula and decodes trajectories with a Hierarchical Key-Value (HKV) Mixture-of-Experts (MoE) Transformer. Formally,
\begin{equation}
    f_\theta:\; \mathcal{I}\times \varphi\times \mathbf{p}_0\times \mathcal{R} \;\mapsto\; \hat{\mathbf{Y}}_{1:T}; 
    \hat{\mathbf{y}}_t=(x_t,y_t,\psi_t)
\end{equation}
where $\mathcal{I}$ is the multi-view concatenated image tensor, $\varphi$ is the STL specification, $\mathbf{p}_0=(x_0,y_0,\psi_0)$ is the starting pose, and $\mathcal{R}$ is the region bounds.

\noindent\textbf{Components.} The \textbf{S-MSP} architecture comprises: (i) an image encoder that yields a compact token memory; (ii) a text encoder over a semantic rendering of $\varphi$; (iii) a pose–region encoder; (iv) a structural STL encoder that parses $\varphi$ into typed tokens aligned with time; and (v) an HKV–MoE trajectory decoder that autoregressively predicts waypoints under multi-source cross-attention.


\noindent\textbf{Baseline.} The baseline we proposed is a plain, non-MoE Transformer, depicted in Fig.~\ref{fig:model} by replacing the red-boxed module with a standard Transformer block. It shares the same I/O interface as our final model: it ingests raw multi-view RGB images and STL task specifications and outputs dynamically feasible trajectories. But it contains no Mixture-of-Experts modules and no STL-conditioned routing

\subsection{Encoders}

\subsubsection{Image Encoder}
We adopt a multi-view image encoder initialized from a pretrained backbone and fine-tuned end-to-end in our pipeline. Given a RGB concatenated image tensor $\mathcal{I}\in\mathbb{R}^{3\times H\times W}$, the encoder outputs a compact token sequence:
\begin{equation}
\mathbf{V} \;=\; \mathrm{ImageEncoder}(\mathcal{I}) \;\in\; \mathbb{R}^{N_v \times d},
\label{eq:image-tokens}
\end{equation}
where $d$ is the embedding dimension and $N_v$ is the post-aggregation token count.

Concretely, we start from a Swin-B backbone (IN22K-ft-1K), take image features from stages 2–4 and align them to a unified resolution (\texttt{out\_indices}=(1,2,3), \texttt{unify\_level}=2). Two cross-attention aggregation layers with learnable latent tokens (512 queries) compress the fused pyramid and cap the key–value memory to $\sim$4K tokens for efficient decoder cross-attention. We add 2D spatial and temporal positional encodings (with optional coordinate channels) before passing $\mathbf{V}$ to the decoder.

\subsubsection{STL Text Encoder}
We encode the formatted STL string into a vector $\mathbf{l}\in\mathbb{R}^{d}$ and map it to
\begin{equation}
\mathbf{L} = \mathrm{Proj}(\mathbf{l}) \in \mathbb{R}^{m_{\text{text}}\times d},
\label{eq:STL}
\end{equation}
where $m_{\text{text}}$ is the number of text-conditioning tokens (we use $m_{\text{text}}=1$ by default). $\mathbf{L}$ is appended to the decoder's cross-attention memory.

\subsubsection{Pose/Region Encoder}
Given the start pose $\mathbf{p}_{0}=(x_{0},y_{0},\psi_{0})$ and the rectangular feasible task region $\mathcal{R}=[x_{\min},y_{\min},x_{\max},y_{\max}]$, the encoder produces
\begin{equation}
\mathbf{c_1, c_2} \;=\; E_{\text{pose/reg}}^{\text{ctx}}(\mathbf{p}_{0},\mathcal{R}) \;\in\; \mathbb{R}^{d};
\mathbf{p}_{0}^{\text{abs}} \;=\; \mathbf{p}_{0} \;\in\; \mathbb{R}^{3}.
\label{eq:pose-region}
\end{equation}
Here $\mathbf{c_i}$ serves as a context vector (concatenated to the decoder’s cross-attention memory), while $\mathbf{p}_{0}^{\text{abs}}$ is passed to the decoder to seed the initial state and anchor the predicted trajectory in the world frame.

\subsubsection{Structural STL Encoder}

We parse the STL specification $\varphi$ into an operator–predicate abstract syntax tree(AST) and linearize it in preorder with a fixed sibling order to obtain a typed token sequence. For each node $m$ we form an embedding by summing token, time-interval, and depth cues:
\begin{subequations}\label{eq:stl-token-enc}
\begin{align}
\mathbf{S} &= [\,\mathbf{s}_1;\ldots;\mathbf{s}_{L}\,] \in \mathbb{R}^{L\times d},\\
\mathbf{s}_m &= E_{\text{tok}}(\mathrm{id}_m)\;+\;E_{\text{interval}}(\tilde{\Delta}_m)\;+\;E_{\text{depth}}(d_m),
\end{align}
\end{subequations}
where $L$ denotes the number of subformulaes, $\mathrm{id}_m$ indexes a unified vocabulary covering operators (\texttt{F/G/and/or}) and targeted region tokens; 
$\tilde{\Delta}_m=[\tilde{a}_m,\tilde{b}_m]\in[0,1]^2$ is the normalized time interval (obtained by dividing by the planning horizon), and $d_m$ is the AST depth. 
We prepend a learnable $\langle\mathrm{CLS}\rangle$ token and feed $[\langle\mathrm{CLS}\rangle;\mathbf{S}]$ to a small Transformer encoder; the $\langle\mathrm{CLS}\rangle$ output serves as the global formula embedding, while per-token outputs are exposed for downstream structural routing.

\subsection{Autoregressive Trajectory Decoder with STL-Structured HKV–MoE}
\label{sec:traj-decoder-hkvmoe}
We decode trajectories autoregressively using a causal Transformer with cross-attention over a shared memory
\[
\mathbf{M}_0 \;=\; \big[\,\mathbf{V}\;;\;\mathbf{L}\;;\;\mathbf{c_1,c_2}\;\big]
\in \mathbb{R}^{(N_v+m_{\text{text}}+2)\times d},
\]
where $\mathbf{V}$ are image tokens (Eq.~\eqref{eq:image-tokens}), $\mathbf{L}\in\mathbb{R}^{m_{\text{text}}\times d}$ are STL text tokens (Eq.~\eqref{eq:STL}) and $\mathbf{c_i}\in\mathbb{R}^{d}$ is the pose/region context (Eq.~\eqref{eq:pose-region}). With $\mathbf{y}_0=\mathbf{p}_0^{\text{abs}}$, the step-$t$ query
\[
\mathbf{q}_t \;=\; E_{\text{frame}}(\mathbf{y}_{t-1}) \;+\; E_{\text{time}}(t) \;+\; E_{\text{pos}}(t)
\]
attends to the autoregressive prefix and $\mathbf{M}_{t-1}$, yielding
\[
\mathbf{h}_t \;=\; \mathrm{DecBlock}(\mathbf{q}_t;\mathbf{M}_{t-1});\quad
\hat{\mathbf{y}}_t \;=\; W_{\text{head}}\,\mathbf{h}_t \in \mathbb{R}^{3}.
\]
During training, schedule sampling with rate $\rho$ replaces $\hat{\mathbf{y}}_t$ by ground-truth position token $\mathbf{y}_t$ with probability $\rho$:
\[
\tilde{\mathbf{y}}_t = \rho\times\mathbf{y}_t + (1 - \rho)\times\hat{\mathbf{y}}_t
\]

The chosen next-frame embedding is appended to update the memory:
\[
\mathbf{M}_{t} \;=\; \big[\,\mathbf{M}_{t-1}\,;\; E_{\text{frame}}(\tilde{\mathbf{y}}_t)+E_{\text{time}}(t)+E_{\text{pos}}(t)\,\big].
\]

\noindent\textbf{STL-Structured MoE.}
Beyond the standard MoE architecture, we develop a \emph{STL-conditioned} expert-routing scheme that harnesses the compositional structure inherent in the temporal logic specification.
$\mathbf{S}$ are structural STL tokens (Eq.~\eqref{eq:stl-token-enc}) denote the per-token structural memory obtained from the structural STL encoder. 
For each decoding step $t$, we determine the \emph{innermost} temporal operator whose bounded interval covers $t$, denoted by $\mathrm{op}(t)$, and assign a coarse time band within that interval, $\mathrm{band}(t)\in\{0,\ldots,K{-}1\}$, where $K$ is the pre-defined interval gap. 
A deterministic map $b(\cdot)$ combines operator family and time band into one of $B$ routing buckets:
\begin{equation}
o_t = b\big(\mathrm{op}(t),\,\mathrm{band}(t)\big)\in\{0,\dots,B{-}1\};t=1,\dots,T
\label{eq:stl-bucket}
\end{equation}
This leverages STL compositionality: formulas decompose into operator-scoped subgoals with finite temporal support, yielding semantically meaningful buckets.

In $L_{\text{moe}}$ decoder layers, the position-wise feed-forward network (FFN) is replaced by a sparse MoE gated \emph{within} the selected bucket $o_t$. 
Let $\{\mathcal{E}_b\}_{b=0}^{B-1}$ be a partition of experts into buckets and $\mathcal{B}(o_t){:=}\mathcal{E}_{o_t}$ the active expert set at step $t$. 
Given hidden state $\mathbf{h}_t$, the MoE output is
\begin{equation}
\mathrm{FFN}^{\text{MoE}}(\mathbf{h}_t; o_t)
\;=\;
\sum_{e \in \mathcal{E}_{\text{top-}k}(o_t)} g_{t,e}\,\mathrm{FFN}_e(\mathbf{h}_t),
\label{eq:moe-sum}
\end{equation}
where $\mathcal{E}_{\text{top-}k}(o_t)\subseteq \mathcal{B}(o_t)$ denotes the top-$k$ experts selected \emph{from the bucket} and the mixture weights $g_{t,e}$ are normalized over $\mathcal{B}(o_t)$ by a lightweight hierarchical key–value router:
\begin{equation}
g_{t,\cdot}
\;=\;
\mathrm{softmax}_{e \in \mathcal{B}(o_t)}
\!\left(
\frac{\langle W_q \mathbf{h}_t,\; \mathbf{K}_e\rangle}{\sqrt{d_k}}
\right),
\sum_{e\in \mathcal{B}(o_t)} g_{t,e}=1.
\label{eq:gating}
\end{equation}
A shared dense pathway runs in parallel and the outputs are blended by a learned gate $\alpha_t\in[0,1]$ (scalar or channel-wise):
\begin{equation}
\mathrm{FFN}_{\text{final}}(\mathbf{h}_t)
\;=\;
\mathrm{FFN}_{\text{shared}}(\mathbf{h}_t)
\;+\;
\alpha_t\,\mathrm{FFN}^{\text{MoE}}(\mathbf{h}_t).
\label{eq:final-ffn}
\end{equation}

This STL-structured routing (i) promotes expert specialization by operator family and coarse time band; (ii) restricts competition to the active bucket, reducing cross-family interference and gradient contention; (iii) enforces sparse activation that allocates capacity where needed, improving sample efficiency; and (iv) yields interpretable routing, as expert activations align with subformulae and their temporal support. 
To prevent expert collapse and maintain utilization, we employ standard auxiliary regularizers (e.g., load-balancing and entropy terms) and recover the dense decoder in the limit $\alpha_t\!\to\!0$.

\subsection{Training Objective}
We minimize a weighted sum of reconstruction, feasibility, obstacle, STL, and MoE regularization terms. For
$\hat{\mathbf{Y}}=(\hat{\mathbf{y}}_{1},\dots,\hat{\mathbf{y}}_{T})$ with $\hat{\mathbf{y}}_{t}=(\hat{x}_{t},\hat{y}_{t},\hat{\psi}_{t})$
and ground truth $\mathbf{Y}=(\mathbf{y}_{1},\dots,\mathbf{y}_{T})$,
\begin{align}\label{eq:total-loss}
\mathcal{L}
\;=\;
w_{\text{rec}}\!\sum_{t=1}^{T}\!\big\|\hat{\mathbf{y}}_{t}-\mathbf{y}_{t}\big\|_2^{2}
\;+\;
w_{\text{end}}\big\|\hat{\mathbf{y}}_{T}^{xy}-\mathbf{y}_{T}^{xy}\big\|_2^{2}
\;
\\ \notag
+\;w_{\text{obs}}\,\mathcal{L}_{\text{obs}}(\hat{\mathbf{Y}})
\;+\;
w_{\text{feas}}\,\mathcal{L}_{\text{feas}}(\hat{\mathbf{Y}})
\;
\\ \notag
+\;w_{\text{stl}}\,[\,\gamma-\rho_{\varphi}(\hat{\mathbf{Y}})\,]_+
\;+\;
w_{\text{moe}}\,\mathcal{R}_{\text{MoE}},
\end{align}
where $\rho_{\varphi}$ is a differentiable STL robustness score (positive when the formula is satisfied), $[\cdot]_+=\max(\cdot,0)$ is a hinge, and $\gamma\!\ge\!0$ is a is the truncation factor, which is used to encourage the policy to improve ``hard'' trajectories (with robustness scores $< \gamma$) rather than further increasing ``easy" trajectories that already achieve high robustness scores ($\geq\gamma$).

Feasibility uses a hinge on step length and heading change:
\begin{align}
\mathcal{L}_{\text{feas}}(\hat{\mathbf{Y}})
\;=\;
\sum_{t=2}^{T}\Big[\,\big\|\hat{\mathbf{y}}_{t}^{xy}-\hat{\mathbf{y}}_{t-1}^{xy}\big\|_2-d_{\max}\,\Big]_+
\;+\;
\\ \notag
\lambda_{\psi}\sum_{t=2}^{T}\Big[\,\big|\hat{\psi}_{t}-\hat{\psi}_{t-1}\big|-\delta_{\max}\,\Big]_+.
\end{align}
Obstacle avoidance applies a smooth barrier/soft-distance surrogate to forbidden regions $\{\mathcal{O}_k\}$:
\begin{align}
\mathcal{L}_{\text{obs}}(\hat{\mathbf{Y}})
\;=\;
\frac{1}{T}\sum_{t=1}^{T}\;\Phi\!\big(\hat{\mathbf{y}}_{t}^{xy};\,\{\mathcal{O}_k\}\big),
\end{align}
with $\Phi$ any differentiable proxy (e.g., softplus of negative signed distance).

MoE regularization discourages expert collapse and promotes balanced routing:
\begin{align}
\mathcal{R}_{\text{MoE}}
\;=\;
\lambda_{\text{bal}}\,\mathcal{L}_{\text{bal}}
\;+\;
\lambda_{\text{ent}}\,\mathcal{L}_{\text{ent}},
\end{align}
where $\mathcal{L}_{\text{bal}}$ encourages uniform expert importance and $\mathcal{L}_{\text{ent}}$ penalizes overly peaky gate distributions; either can be instantiated with standard choices.

Schedules are kept simple and effective: the teacher-forcing probability decays linearly
$\rho(e)=\max\{0,\,1-e/E\}$ over epochs $e=0,\dots,E$, and secondary weights are warmed up
$w_j(e)=r(e)\,w_j^\star$ with $r(e)$ increasing from $r_{\min}\!\in(0,1]$ to $1$ over the first $E_{\text{warm}}$ epochs for $j\in\{\text{obs},\text{feas},\text{stl},\text{moe}\}$. The MoE blend is initialized toward the dense path to recover the non-MoE decoder at early training and is learned end-to-end thereafter.

\begin{algorithm}[t]
\caption{STL-Aware RRT-Guided Local Replanning (Triggered Segment Repair, \textbf{TSP})}
\label{alg:stl-rrt-repair}
\DontPrintSemicolon
\SetKwInOut{KwIn}{Input}\SetKwInOut{KwOut}{Output}
\KwIn{Nominal trajectory $\hat{\mathbf{Y}}=\{\hat{y}_1,\dots,\hat{y}_T\}$; STL formula $\varphi$; horizon $H$; goal-bias $\beta$; sampling times $M$; limits $(d_{\max},\Delta\theta_{\max})$}
\KwOut{Repaired trajectory $\widetilde{\mathbf{Y}}$}

$\widetilde{\mathbf{Y}}\leftarrow\hat{\mathbf{Y}}$;\quad $t\leftarrow 1$\;
\While{$t \le T$}{
  $H_t=\min(H,\,T-t-1)$\;
  \If{$CK(\widetilde{\mathbf{Y}},t)$ \textbf{or} $\rho(\widetilde{\mathbf{Y}}_{t:t+H_t},\,\varphi) < 0$}{
    Initialize tree $\mathcal{T}$ with root $z_{0} = \widetilde{y}_{t-1}$; set $z_{H_t+1} = \widetilde{y}_{t+H_t+1}$\;

    \For{$m=1$ \KwTo $M$}{
      Sample target $q$: with prob $\beta$ sample near anchors $\{\hat{p}_\tau,\dots,\hat{p}_{\tau+H_t}\}$; else sample uniformly\;
      $z_{\text{near}}\leftarrow\texttt{nearest}(\mathcal{T},q)$\;
      $z_{\text{new}}\leftarrow\texttt{run}(z_{\text{near}},q;d_{\max},\Delta\theta_{\max})$\;

    insert $z_{\text{new}}$ and edge $(z_{\text{near}},z_{\text{new}})$ into $\mathcal{T}$\;
    }
    $\mathbf{Y}_{new} = \texttt{path}(\mathcal{T}, z),$ with min $\mathcal{J}$ via \eqref{eq:stl-seg-cost}, $\forall z$ satisfies $\texttt{height(z)} = H_t$ \textbf{and} $feasible(z, z_{H_t + 1})$
  }

  $t \leftarrow t+1$\;
}
\Return $\widetilde{\mathbf{Y}}$\;
\end{algorithm}

\subsection{RRT-Guided Local Replanning with STL Awareness}
Given a nominal trajectory $\hat{\mathbf{Y}}=\{(x_t,y_t,\theta_t)\}_{t=1}^{T}$, we run a rollout-and-check loop.
If a forward lookahead at time $\tau$ indicates (i) collision/kinematic violation, denoted by $CK(Y, \tau)$, or (ii) \emph{STL robustness}
of the local window is negative, we trigger short-horizon RRT-style replanning on $[\tau,\tau{+}H]$.
Expansion is goal-biased: with probability $\beta$ targets are sampled near nominal anchors
$\{\hat{p}_t\}_{t=\tau}^{\tau+H}$,
otherwise uniformly. Candidate edges must satisfy local limits
$\|p_t-p_{t-1}\|\!\le\!d_{\max}$ and $|\theta_t-\theta_{t-1}|\!\le\!\Delta\theta_{\max}$.
Among feasible branches that reach $\tau{+}H$, we select the path minimizing
\begin{equation}
\label{eq:stl-seg-cost}
\mathcal{J}
= \sum_{t=\tau}^{\tau+H}\!\Big(\lambda_1\|p_t-\hat{p}_t\|_2^2+\lambda_2\,\kappa_t^2\Big)
\;-\; \lambda_3\,\tilde{\rho}_{\varphi}\!\big(\widetilde{\mathbf{Y}}_{\tau:\tau+H}\big),
\end{equation}
where $p_t=(x_t,y_t)$, $\kappa_t$ is a discrete curvature proxy, and $\tilde{\rho}_{\varphi}$ is the STL robustness of the segment (higher is better). The repaired segment is optionally smoothed (e.g., cubic spline)
and spliced back into $\hat{\mathbf{Y}}$. The full procedures, termed \textbf{TSP}, is shown in Algorithm~\ref{alg:stl-rrt-repair}. 

\begin{table*}[!t]
  \centering
  \setlength{\tabcolsep}{6pt}
  \renewcommand{\arraystretch}{1.15}
  \resizebox{\textwidth}{!}{
  \begin{tabular}{lccccccccc}
    \toprule
    Method & single-F & single-G & only-AND & only-OR & only(AND--OR) & All (ID) & OOD-1 (All) & OOD-2 (All) & OOD-3 (All) \\
    \midrule
    Baseline (no-MoE)        & 72.75 & 60.75 & 61.25 & 64.88 & 60.50 & 61.50 & 7.50 & 8.62 & 12.62 \\
    Ours(S-MSP)      & 77.75 & 77.00 & 63.50 & 79.00 & 62.50 & 71.00 & 7.88 & 6.67 & 14.00 \\
    Baseline + TSP & \textbf{90.88} & 79.00 & 77.75 & 82.25 & 74.33 & 77.88 & 12.50 & \textbf{12.88} & 19.25 \\
    Ours(S-MSP) + TSP            & 89.00 & \textbf{90.00} & \textbf{85.83} & \textbf{93.00} & \textbf{84.50} & \textbf{88.00} & \textbf{13.67} & 10.82 & \textbf{23.00} \\
    \bottomrule
  \end{tabular}}
  \caption{Success rate on ID subsets and aggregated OOD scenes.}
  \label{tab:results}
\end{table*}

\section{Experiments}\label{sec:case}
\subsection{Dataset and Metrics}

\noindent\textbf{Dataset.}
We construct a high-fidelity, Gazebo-based benchmark for factory–logistics navigation under Signal Temporal Logic (STL) guidance. Each episode includes synchronized multi-view RGB streams with calibrated intrinsics/extrinsics, ego states, polygonal maps of static obstacles and task regions, a formally annotated STL specification from a bounded-horizon fragment, and an expert reference trajectory. The benchmark covers representative warehouse layouts (aisles, racks, pallets) with domain randomization in geometry, placement, lighting, and textures. To assess generalization, we adopt two regimes: \emph{ID-layout with novel STL tasks} (ID-L/Novel-$\Phi$), where layouts are seen during training but task specifications are new, and \emph{OOD-layout with novel STL tasks} (OOD-L/Novel-$\Phi$), where both layouts and task compositions are unseen. Splits are disjoint by layout and STL templates. We standardize a fixed prediction horizon and provide per-episode metadata for reproducibility.

\noindent\textbf{Metrics.}
We evaluate predicted trajectories along only one ax: \emph{Success Rate (SR)} for episodes that satisfy the stl task and remain dynamically feasible. Unless otherwise specified, we report per-regime episode-level point estimates for ID-L/Novel-$\Phi$ and OOD-L/Novel-$\Phi$, together with scenario-wise breakdowns.

\subsection{Implementation details}
We train on a single NVIDIA A800 (80\,GB) with batch size $12$ for $24$ epochs using AdamW (betas $0.9/0.95$) and three parameter groups: (i) weight-decayed ($10^{-2}$), (ii) norm/bias/embedding without decay, and (iii) router centroids (\texttt{expert\_key}) with a $2\times$ larger learning rate and a small decay $10^{-3}$; the base learning rate is $1{\times}10^{-5}$. Unless noted, we set $d{=}512$, horizon $T{=}80$, and a $4$-layer decoder with $4$ heads and hidden width $128$; MoE is enabled in the last $2$ layers with $B{=}6$ buckets, $E'{=}2$ experts per bucket, temperature $\tau{=}1.0$, and in-bucket top-$k{=}2$ (dropout $0.1$). What's more, the sampling probability $\beta$ in \textbf{TSP} is 0.5. The source codes are available at~\url{https://github.com/yc7421cmd/E2E-Signal-Temporal-Logic-Planner.git}.

\subsection{Ablation Study}
Table~\ref{tab:scaling-law} reports All(ID) ablations on three factors while keeping splits, schedule, and optimiser settings fixed: (a) the \emph{loss margin} $\gamma$ in the STL hinge term of Eq.~\eqref{eq:total-loss}, (b) the \emph{sharpness} $k$ of the smooth $\max/\min$ used in robustness aggregation, and (c) the image resolution/backbone.

\noindent\textbf{Loss margin.}
$\gamma$ enters only through the hinge $[\,\gamma-\rho_{\varphi}(\hat{\mathbf{Y}})\,]_+$. Small margins under-penalize near-violations and admit borderline solutions; overly large margins keep the hinge active for most samples, slowing optimization and trading off with feasibility/collision terms. We observe a wide stable regime for moderate margins (e.g., $\gamma\!\in\![0.1,0.5]$); pushing $\gamma$ higher yields diminishing or negative returns. We therefore adopt $\gamma{=}0.2$ in all main results.

\noindent\textbf{Smooth-aggregation sharpness.}
$k$ controls the tightness of the differentiable $\widetilde{\max}/\widetilde{\min}$ used to approximate formulae robust. Very small $k$ over-smooths and biases robustness; very large $k$ approaches hard max/min and produces sparse, ill-conditioned gradients. A mid-range setting achieves the best bias–variance trade-off; we use $k{=}300$ by default.

\noindent\textbf{Image resolution and backbone.}
Under an iso-compute setting, increasing input resolution improves geometric fidelity and adherence to fine spatial constraints, but exhibits clear diminishing returns alongside higher memory/FLOPs. Within Swin-B, using larger attention windows (w12) at medium–high resolutions expands the effective receptive field without increasing the number of encoder tokens. In our ablations, scaling from 320$\times$480 to 480$\times$720 delivers the most pronounced gains, while a further jump to 640$\times$960 yields only marginal improvement at disproportionate cost. We therefore adopt w12 at a medium–high resolution (e.g., 512$\times$768) as the default.

\begin{table}[t]
  \centering
  \begin{minipage}{0.49\linewidth}
    \centering
    a) Satisfication threshold $\gamma$ \\
    \resizebox{1\textwidth}{!}{
      \begin{tabular}{ccc}
        \hline
        $\gamma$ & Val Acc. & Test Acc. \\ \hline 
        0.0  & 66.00 & 66.67 \\
        0.1  & 71.50 & 69.50 \\ 
        0.2  & 73.50 & 71.00 \\ 
        0.5  & 71.00 & 68.50 \\ 
        0.8  & 66.75 & 65.00 \\ 
        1.0  & 61.67 & 62.00 \\ \hline
      \end{tabular}
    }
  \end{minipage}
  \hfill
  \begin{minipage}{0.49\linewidth}
    \centering
    b) Sharpness $k$ for $\widetilde{\max}/\widetilde{\min}$ \\
    \resizebox{1\textwidth}{!}{
      \begin{tabular}{ccc}
        \hline
        $k$ & Val Acc. & Test Acc. \\ \hline 
        1  & 55.33 & 55.00 \\
        5  & 53.00 & 53.00 \\ 
        10  & 63.33 & 63.00 \\ 
        100  & 64.88 & 65.00 \\ 
        1000  & 70.75 & 69.50 \\ 
        10000  & 69.00 & 68.00 \\  \hline
      \end{tabular}
    }
  \end{minipage}
  \\
  \vspace{1em}
  \begin{minipage}{0.88\linewidth}
    \centering
    c) Image resolution and backbone \\
    \resizebox{1\textwidth}{!}{
      \begin{tabular}{lcccc}
        \hline
        Model & Resolution. & Window. & Val Acc. & Test Acc. \\ \hline 
        Swin-B w7 & 320$\times$480 & 7 & 63.00 & 61.50 \\ 
        Swin-B w7 & 480$\times$720 & 7 & 66.33 & 66.00 \\ 
        Swin-B w7 & 640$\times$960 & 7 & 67.50 & 65.00 \\ 
        Swin-B w12 & 384$\times$576 & 12 & 66.00 & 68.75 \\ 
        Swin-B w12 & 512$\times$768 & 12 & 73.50 & 71.00 \\ 
        \hline
      \end{tabular}
    }
  \end{minipage}
  \caption{Different setups in training S-MSP.}\label{tab:scaling-law} 
\end{table}

\subsection{Main Results}
We evaluate on five in-distribution subsets, \texttt{single-F}, \texttt{single-G}, \texttt{only-AND}, \texttt{only-OR}, \texttt{only(AND-OR)} and their union set \texttt{All (ID)}, and three aggregated out-of-distribution (OOD) suites (\texttt{OOD-1/2/3 (All)}). The primary metric is \emph{success rate}, defined as jointly satisfying the STL specification, remaining collision-free, and meeting kinematic feasibility. All entries in Table~\ref{tab:results} are percentages.

\begin{figure*}[!t]
  \centering
    \subfloat[ID]{\includegraphics[width=0.24\textwidth,trim={130 20 28 20},clip]{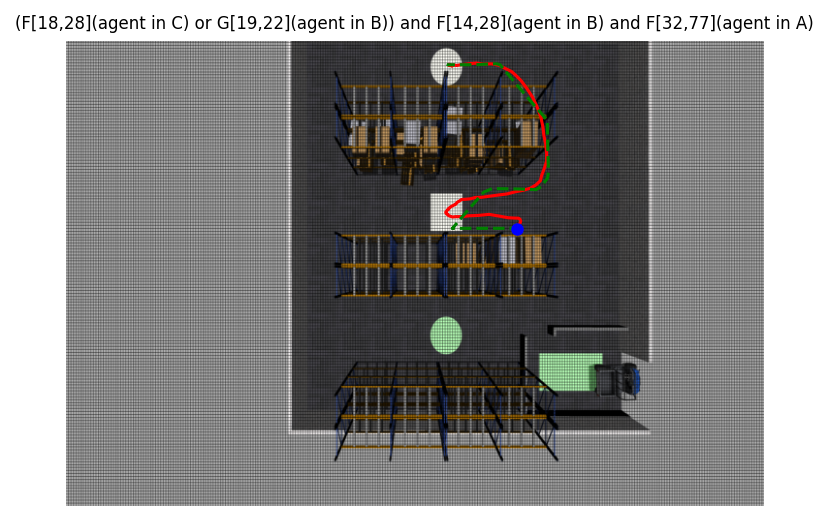}}\hfill
    \subfloat[OOD-1]{\includegraphics[width=0.24\textwidth,trim={132 20 25 20},clip]{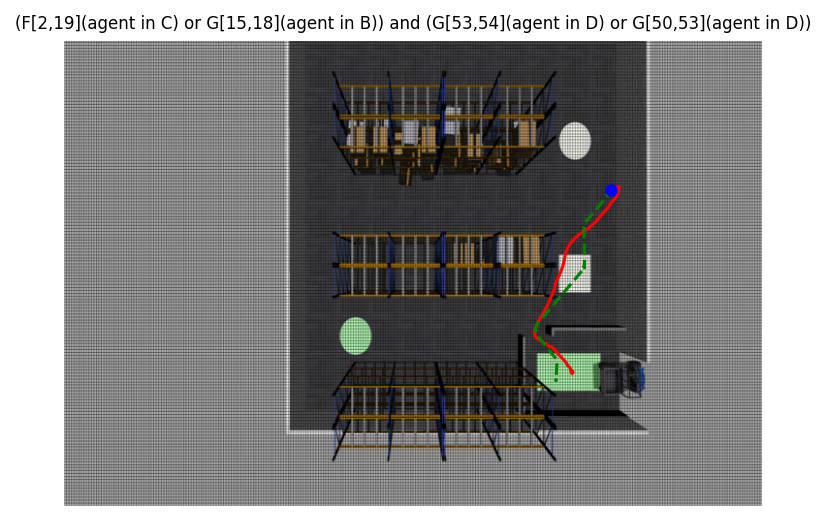}}\hfill
    \subfloat[OOD-2]{\includegraphics[width=0.24\textwidth,trim={178 20 83 20},clip]{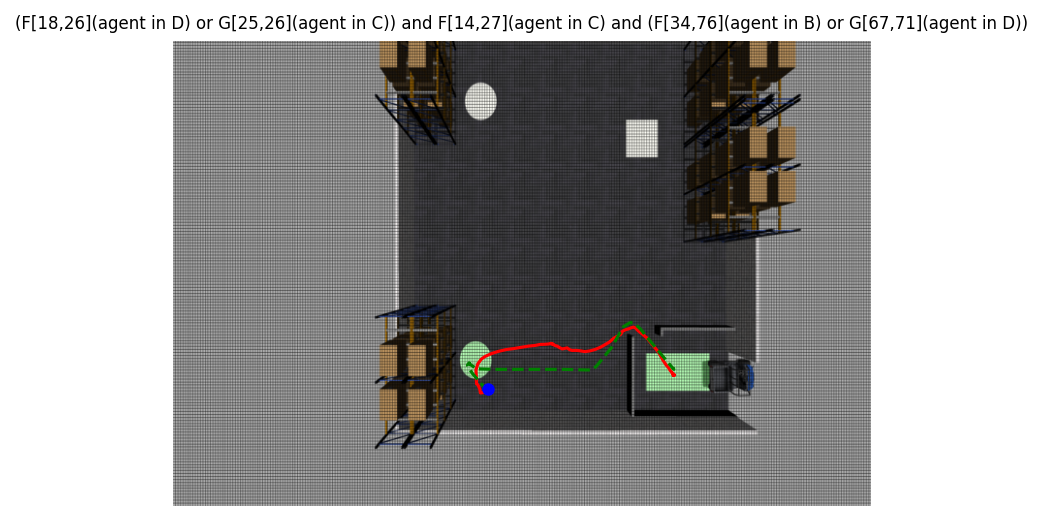}}\hfill
    \subfloat[OOD-3]{\includegraphics[width=0.24\textwidth,trim={102 20 07 20},clip]{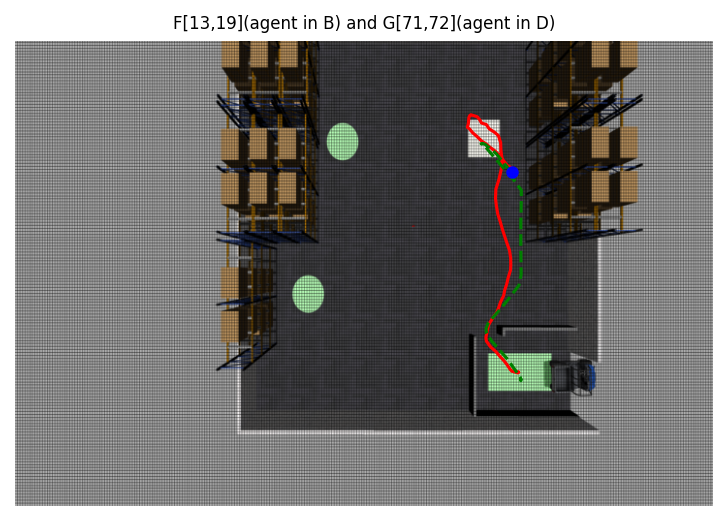}}
  \caption{Qualitative rollouts on representative tasks from ID and OOD suites.}
  \label{fig:qual_rollouts}
\end{figure*}

Overall, the \textbf{Baseline (no-MoE)} shows limited generalization, particularly on composite formulas (notably \texttt{only(AND-OR)}) and under OOD shift. A post-hoc \textbf{TSP} improves feasibility and safety but yields only modest gains in STL satisfaction, since downstream filtering cannot correct upstream semantic planning errors. Relative to a simple transformer baseline, \textbf{S-MSP} uses structured MoE routing aligned with STL operator families and temporal segments, reducing upstream semantic mismatches and improving reliability under shift. Quantitatively (Table~\ref{tab:results}), on ID it lifts “All (ID)” from $61.50\%$ to $71.00\%$ $(+9.50\,\mathrm{pp})$, with pronounced gains on \texttt{single-G} $(+16.25\,\mathrm{pp})$ and \texttt{only-OR} $(+14.12\,\mathrm{pp})$. With the \textbf{TSP} post-processor, the advantage amplifies: “All (ID)” reaches $88.00\%$ versus $77.88\%$ $(+10.12\,\mathrm{pp})$, and \texttt{only(AND-OR)} improves by $+10.17\,\mathrm{pp}$. Under OOD shift, S-MSP$+$TSP attains the best mean across suites $(15.83\%\ \mathrm{vs.}\ 14.88\%)$, led by \texttt{OOD-3} $(+3.75\,\mathrm{pp})$, and achieves best-in-column on $5/6$ ID subsets (plus the ID aggregate) and $2/3$ OOD suites. Here, pp denotes percentage points.

To better illustrate the behavior of our planner, we further include four full-width qualitative rollouts—one each from the ID, OOD-1, OOD-2, and OOD-3 suites, showing successful trajectories that satisfy the STL specification while remaining collision-free and kinematically feasible in Fig.~\ref{fig:qual_rollouts}. We use a factory-transport environment with four zones, white circle ($A$), white square ($B$), green circle ($C$), green rectangle ($D$), and static shelves/walls delimiting the drivable free space $\mathcal{S}$.
The planner is trained on the ID layout and evaluated (without finetuning) on one ID and three OOD layouts. Blue point marks the start, green is the expert trajectory, red is the planned trajectory and $v$ means vehicle. Four STL tasks are showing below:
\begin{subequations}\label{eq:struct-enc}
\begin{align}
\phi_a = & (\eventually[18,28](v\in C) \vee \always[19,22](v\in B)) \wedge \\ \notag
& \eventually[14,28](v\in B) \wedge\eventually[32,77](v\in A);\\
\phi_b = &(\eventually[2,19](v\in C)\vee\always[15,18](v\in B))\wedge\\\notag
&(\always[53,54](v\in D)\vee \always[50,53](v\in D));\\
\phi_c = & (\eventually[18,26](v\in D)\vee\always[25,26](v\in C))\wedge\\\notag
&(\eventually[34,76](v\in B)\vee\always[67,71](v\in D))\wedge\\\notag
&\eventually[14,27](v\in C);\\
\phi_d = & \eventually[13,19](v\in B)\wedge\always[71,72](v\in D).
\end{align}
\end{subequations}

The rollouts provide evidence that the model has learned \emph{image- and task-conditioned} planning, producing collision-free and kinematically admissible trajectories across both ID and OOD layouts. Given only multi-iew image observations and finite-horizon STL specifications, the planner consistently satisfies the prescribed reach/sequence predicates while remaining within the drivable set and respecting per-step motion limits, without per-layout tuning. This indicates transfer of both scene understanding and symbolic goal structure rather than memorization of a fixed particular map.

Such behavior confers several system-level benefits: (i) \emph{compositionality}, as STL-defined objectives can be recombined without retraining; (ii) \emph{robustness to moderate distribution shift}, as performance persists under shelf/aisle perturbations; (iii) \emph{verifiability}, since satisfaction of $\mathbf{F}/\mathbf{G}$ predicates and feasibility constraints yields a principled certificate of correctness; (iv) \emph{operational simplicity}, reducing hand-engineered interfaces between perception and planning by training end-to-end; and (v) \emph{diagnosability}, as robustness margins and violation sets localize potential failure modes for targeted remediation. Collectively, these properties suggest a practical path toward deployable, specification-driven planning in structured industrial environments.

\section{Conclusion}
In this work, we presented, to the best of our knowledge, the first end-to-end model \textbf{S-MSP} for STL-specified robotic control, mapping synchronized sensory inputs and symbolic specifications directly to feasible trajectories. A structure-aware sparse Mixture-of-Experts exploits STL compositionality and temporal scope, yielding faster convergence and high satisfaction and feasibility at low latency. For deployability, we augment the policy with a runtime repair layer that invokes an RRT-style local replanner to modify only violating subsegments while respecting dynamics and obstacles. The experimental results validate the effectiveness and robustness of the proposed approach.

In this work, we restrict attention to STL fragments without nested temporal operators. Future work will extend the framework to nested operators and incorporate reinforcement learning for preference alignment and adaptation under distributional shifts to improve deployment robustness.

\bibliographystyle{abbrv}
\bibliography{reference} 
\end{document}